\documentclass[letterpaper, 10 pt, conference]{ieeeconf}  

\IEEEoverridecommandlockouts                              

\usepackage{csquotes}
\usepackage[capitalize]{cleveref}
\usepackage{graphicx}
\usepackage{csquotes}
\usepackage{tabularray}
\usepackage{tabularx}
\usepackage[T1]{fontenc}
\usepackage[hyphens]{url}
\usepackage{cite}
\let\labelindent\relax
\usepackage{enumitem}

\let\originalparagraph\paragraph
\renewcommand{\paragraph}[1]{\originalparagraph{\textbf{#1}}}

\newcommand*{\subparagraph}[1]{\par\textit{#1}}

\title{\LARGE \bf
How Could Generative AI Support Compliance with the EU AI Act?
\\A Review for Safe Automated Driving Perception
}

\author{Mert Keser$^{1,2}$, Youssef Shoeb$^{2,3}$, and Alois Knoll$^{1}$
\thanks{*The review was written in the context of the "NXT GEN AI Methods" research project funded by  
the German Federal Ministry for Economic Affairs and Climate Action (BMWK), The authors would like to thank the consortium for the successful cooperation}
\thanks{$^{1}$Technical University of Munich, Germany {\tt\small mert.keser@tum.de}}%
\thanks{$^{2}$Continental AG, Germany}%
\thanks{$^{3}$Technical University of Berlin, Germany}%
}

\begin{document}

\maketitle
\thispagestyle{empty}
\pagestyle{empty}

\begin{abstract}

Deep Neural Networks (DNNs) have become central for the perception functions of autonomous vehicles, substantially enhancing their ability to understand and interpret the environment.
However, these systems exhibit inherent limitations such as brittleness, opacity, and unpredictable behavior in out-of-distribution scenarios.
The European Union (EU) Artificial Intelligence (AI) Act, as a pioneering legislative framework, aims to address these challenges by establishing stringent norms and standards for AI systems, including those used in autonomous driving (AD), which are categorized as high-risk AI.
In this work, we explore how the newly available generative AI models can potentially support addressing upcoming regulatory requirements in AD perception, particularly with respect to safety. This short review paper summarizes the requirements arising from the EU AI Act regarding DNN-based perception systems and systematically categorizes existing generative AI applications in AD. 
While generative AI models show promise in addressing some of the EU AI Act's requirements, such as transparency and robustness, this review examines their potential benefits and discusses how developers could leverage these methods to enhance compliance with the Act. The paper also highlights areas where further research is needed to ensure reliable and safe integration of these technologies.

\end{abstract}

\section{Introduction}
\label{sec:intro}

DNNs have revolutionized AD by improving vehicle perception, trajectory prediction, and environmental understanding through advanced sensor data processing \cite{grigorescu2020survey,yurtsever2020survey,velasco2020autonomous}. However, significant challenges remain, particularly in generalization and interpretability, which are critical for safe deployment. As statistical models, DNNs inherently struggle to adapt to novel scenarios that deviate from their training data \cite{keser2023interpretable}. In dynamic environments, unexpected domain shifts such as changing weather conditions, new behaviors, and unfamiliar objects can lead to safety-critical errors \cite{ponn2020identification}. Additionally, the black-box nature of DNNs complicates the interpretation and debugging of their outputs, making it difficult to enhance performance in untested scenarios. Addressing these limitations is essential to ensure the reliability and safety of AD systems.

The term generative AI origins from statistical modelling, where a generative model is one that---other than a discriminative one---allows to generate \emph{new} instances of an observable distribution (e.g., real-world images), possibly conditioned on some target attributes (e.g., a prompt). 
We experienced an increasing influence of Large Language Models (LLMs) \cite{ brown2020language, meta-llama3-webpage} and diffusion models \cite{yang2023diffusion} in various domains. This lead to a notable surge in research exploring so-called generative AI applications, also in AD \cite{huang2023applications}.

As generative AI technologies evolve and find applications in critical systems such as AD, they intersect with emerging regulatory frameworks, such as the EU AI Act. This raises the question: \emph{Can the newly available technologies assist in achieving compliance with upcoming regulations?} For this, a requirement-based approach is particularly vital for developers to understand how generative AI can help to meet the EU AI Act's \cite{euaiactcompromise2024} requirements. While existing surveys have focused on multimodal LLMs \cite{cui2024survey, zhou2023vision, yang2023llm4drive, fu2024drive} and foundational models \cite{huang2023applications} in AD, our survey specifically addresses how generative AI aligns with the EU AI Act's mandates \cite{euaiactcompromise2024}, which will become mandatory for all AI risk categories after 2027. Focusing on research published between 2022 and July 2024 and using keywords such as "transparency," "robustness," "world model," and "visual-question answering," we evaluated recent advancements in generative AI as indexed by Google Scholar. Studies relevant to AD perception systems were included, employing bibliographic snowballing for comprehensive coverage.

The main contributions of this work are:
\begin{itemize}
\item We introduce a systematic \emph{framework to categorize generative AI} applications in AD, particularly focusing on perception safety
\item We provide an outlook on how to align the advancements in generative AI applications in AD, specifically in the context of perception, with the needs arising from regulatory requirements of the EU AI Act.
\end{itemize}

This integrated approach provides a foundational understanding of how generative AI can enhance perception safety in AD while adhering to legislative requirements.

\section{BACKGROUND}
\label{sec:back}

\subsection{Generative AI and Autonomous Driving}
\label{sec:background.GenAIAD}

Generative AI has emerged as a significant field of research, with applications such as image generators \cite{ramesh2022hierarchical} and language models \cite{OpenAIChatGPT} gaining widespread acceptance. The release of ChatGPT \cite{OpenAIChatGPT} has brought these models into the public spotlight, highlighting their potential impact across various domains, including AD.

\paragraph{Image Generation}
In image generation, prominent techniques include Generative Adversarial Networks (GANs) \cite{goodfellow2020generative}, Variational Autoencoders (VAEs) \cite{kingma2013auto}, and Diffusion Models \cite{yang2023diffusion}. Diffusion Models, particularly Stable Diffusion \cite{Rombach_2022_CVPR}, have gained popularity for their ability to generate high-quality images from textual descriptions.


\paragraph{Text Generation}
Text generation has seen a progression from Recurrent Neural Networks (RNNs) \cite{sutskever2011generating} and Long Short-Term Memory Networks (LSTMs) \cite{hochreiter1997long} to Transformer-based models \cite{vaswani2017attention}. LLMs like GPT-3 \cite{brown2020language}, and LLaMA 3 \cite{meta-llama3-webpage} have demonstrated impressive capabilities in natural-language processing. 



\section{Generative Models in Autonomous Driving}
\label{sec:GenerativeModelsinAD}

\begin{figure}[ht]
\vspace{-1em} 
\centering
\includegraphics[width=0.9\columnwidth]{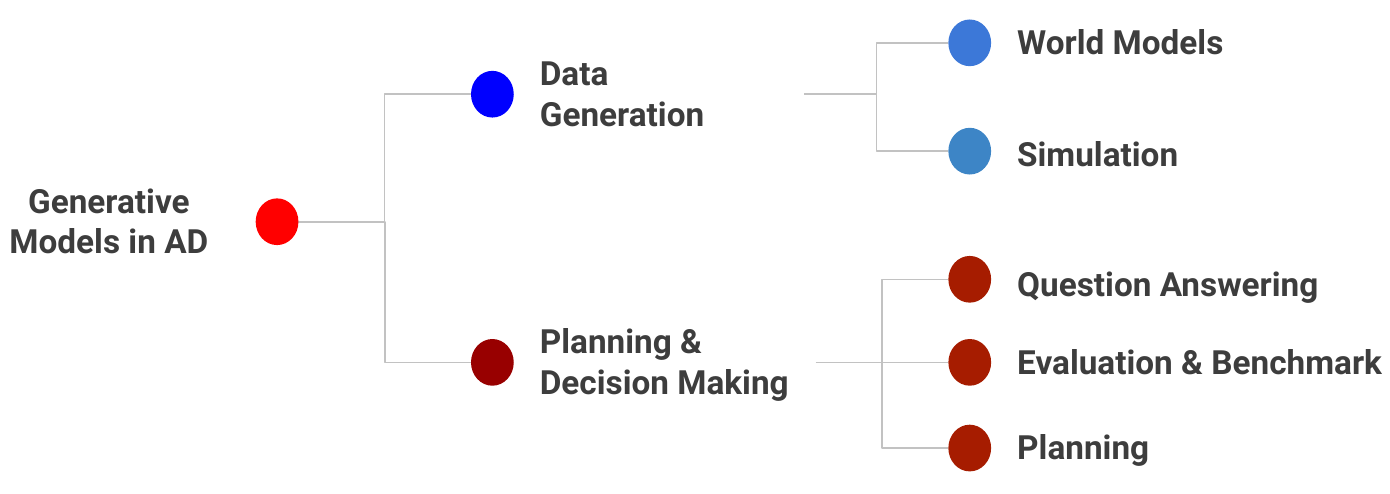}
\vspace{-0.5em} 
\caption{Categorisation of Generative Models in Autonomous Driving}
\label{fig:GenAIinAD}
\vspace{-1.5em} 
\end{figure}

In this section, we categorize generative models for AD based on their application purposes, as shown in \cref{fig:GenAIinAD}.

\subsection{Data Generation}
\label{sec:GenModelsDataGeneration}

Generative models in AD focus on data generation, utilizing both \textit{world models} and \textit{simulation methods} to enhance system training and testing. World models simulate realistic driving scenarios, including rare ‘long-tail’ situations, with photorealistic fidelity and adherence to physical laws. They generate dynamic and static environments, enhancing safety and efficiency by providing scenarios for autonomous systems to learn from \cite{hu2023gaia,wang2024worlddreamer,videoworldsimulators2024}. Inputs like images, prompts, and actions are transformed into detailed outputs such as traffic scenario videos \cite{hu2023gaia,wang2024worlddreamer,videoworldsimulators2024,min2023uniworld,jia2023adriver,jin2023surrealdriver,zhang2023trafficbots,wang2023drivedreamer}. Simulation methods complement world models by creating photorealistic static and dynamic traffic objects, generating safety-critical scenarios, and supporting domain generalization \cite{marathe2023wedge,guo2023controllable,boreiko2023identifying,liudiffusion}. Together, these methods offer a comprehensive approach to modeling complex driving environments, enabling autonomous vehicles to handle diverse and challenging situations.

\begin{table*}
\centering
\renewcommand{\arraystretch}{1.5} 
\caption{Overview of World Models for  Perception of Automated Driving (see \cref{sec:GenModelsDataGeneration})}
\vspace{-1em} 
\label{table:World_Models}
\scriptsize 
\begin{tabularx}{\textwidth}{ccccX}
\hline
\textbf{Model} & \textbf{Year} & \textbf{Input Format} & \textbf{Generated Output} & \multicolumn{1}{c}{\textbf{Model Description}} \\ \hline
MILE \cite{hu2022model} & 2022 & Image & Driving Policy & Prob. Gen. model-based imitation learning approach
for policy learning in AD. \\
SEM2 \cite{wang2023driving} & 2022 & Image & Driving Policy & Multimodal Semantic-Mask based learning of the world and produce driving policy. \\
Drive-WM \cite{wang2023driving} & 2023 & Action-Image-Prompt & Video & Multimodal End-to-end automated driving world model. \\
TrafficBots \cite{zhang2023trafficbots} & 2023 & Image & Video & Uses motion prediction and positional encoding to simulate realistic urban traffic. \\
ADriver-i \cite{jia2023adriver} & 2023 & Image-Action & Video & Multimodal LLMs and video diffusion models to predict control signals in frames. \\
UniWorld \cite{min2023uniworld} & 2023 & Image-Lidar & Video & 4D occupancy maps to understand the environment and predict participant behavior. \\
GAIA-1 \cite{hu2023gaia} & 2023 & Video-Prompt & Video & Realistic traffic scenarios with safety risks and CLIP to align images and text. \\
DriveDreamer \cite{wang2023drivedreamer} & 2023 & Video-Prompt-Action & Video & Text, an image, an HD map, and 3D boxes to generate videos with driving policies \\
CTG++ \cite{zhong2023language} & 2023 & Prompt & Video & Driving scenarios by transforming prompts into loss functions to find trajectories. \\
MUVO \cite{bogdoll2023muvo} & 2023 & Camera-Lidar & Camera-Lidar & Used world model to predict 3D occupancy and camera and lidar observations. \\
OccWorld \cite{zheng2023occworld} & 2023 & Image & Video & 3D occupancy space using a scene tokenizer to understand the surroundings. \\
World Dreamer \cite{wang2024worlddreamer} & 2024 & Action-Prompt-Image & Video & Masked tokens and a spatial temporal transformer for motion and physics in video. \\
Sora \cite{videoworldsimulators2024} & 2024 & Prompt-Image-Video & Image-Video & Generates videos of variable duration and resolution by training diffusion models. \\
Genie \cite{li2024think2drive} & 2024 & Prompt-Sketch-Image-Action & Image-Video & Trained in unsupervised, unlabeled videos to gen. interactive environments. \\
\end{tabularx}
\vspace{-1em} 
\end{table*}

\begin{table*}
\centering
\renewcommand{\arraystretch}{1.5} 
\caption{Overview of Generative Models for Data Creation in AD (see \cref{sec:GenModelsDataGeneration})}
\vspace{-1em} 
\label{table:Simulation_Methods}
\scriptsize
\begin{tabularx}{\textwidth}{ccccX}
\hline
\textbf{Model} & \textbf{Year} & \textbf{Input Format} & \textbf{Generated Output} & \multicolumn{1}{c}{\textbf{Model Description}} \\ \hline
PromptFormer \cite{gong2023prompting} & 2023 & Image & Image & Utilizing generative models for robust semantic segmentation. \\
ContDiff \cite{guo2023controllable} & 2023 & Image-Bbox & Image & Controllable diffusion model for safety-critical scenario generation. \\
BRAVO \cite{loiseau2023reliability} & 2023 & Image & Image & Eval. the robustness of semantic seg. models with images inpainted by diffusion models. \\
SCROD \cite{boreiko2023identifying} & 2023 & Image & Image & Identify systematic errors in object detection with diffusion generated images. \\
WEDGE \cite{marathe2023wedge} & 2023 & Prompt & Image & Synthetic data generation for different weather conditions. \\
DriveSceneGen \cite{sun2023drivescenegen} & 2023 & Image & Image & Generates real-world driving scenarios from scratch using real-world driving datasets. \\
MagicDrive \cite{gao2023magicdrive} & 2023 & Image-B.Box-Pose & Image & Generates multi-camera scenes using 3D annotations, object boxes, and camera parameters \\
Driving Diffusion \cite{li2023drivingdiffusion} & 2023 & Image-Prompt & Video & Use 3D layout to create realistic multi-perspective videos \\
DIDEX \cite{niemeijer2024generalization} & 2024 & Image-Prompt & Image & Domain generalization with diffusion models for synthetic-to real domain adaptation \\
DiffOOD \cite{liudiffusion} & 2024 & Image & Image & Diffusion based out-of-distribution data generation framework. \\

\end{tabularx}
\vspace{-1em} 
\end{table*}

\subsection{Planning \&  Decision Making}
\label{sec:GenModelsDataDecisionMaking}

LLMs enhance cognitive and reasoning capacities in open-world scenarios. Integrated with visual encoders, they are effectively used in AD tasks, offering potential for interpretable AD systems. This subsection covers three main parts: Visual Question Answering (VQA) (\ref{sec:QuestionAnswering}), Evaluation \& Benchmark (\ref{sec:EvaluationBenchmark}), and Planning (\ref{sec:Planning}).

\subsubsection{Question Answering}
\label{sec:QuestionAnswering}


LLMs, with their advanced reasoning and cognitive capacities, are used to navigate complex traffic scenarios. Research integrates chain-of-thought reasoning into multimodal LLMs, simulating step-by-step processes to improve on cognitive tasks \cite{ma2023dolphins, sima2023drivelm, mao2023language, tian2024drivevlm, yuan2024rag, pan2024vlp}. Memory functions enable models to retain information over longer interactions, providing contextual continuity \cite{mao2023language, zhou2024embodied, yuan2024rag, pan2024vlp}. Table \ref{table:VQAandPlanning} summarizes models for VQA and planning tasks, including generating descriptions of traffic scenarios and performing planning tasks for AD \cite{wang2023drivemlm, xu2023drivegpt4, mao2023language, tian2024drivevlm, han2024dme, zhou2024embodied, ding2024holistic}.


\subsubsection{Evaluation \& Benchmark}
\label{sec:EvaluationBenchmark}


Frameworks and datasets evaluate the visual explanations provided by multimodal LLMs. Table \ref{table:VQAandPlanning} summarizes models for VQA and planning tasks. Some frameworks test VQA responses \cite{marcu2023lingoqa,yang2023lidar,qian2024nuscenes,malla2023drama, sima2023drivelm,park2024vlaad,inoue2024nuscenes}, others assess reasoning capacity \cite{marcu2023lingoqa,nie2023reason2drive,malla2023drama,sima2023drivelm}, and some evaluate scene understanding \cite{yang2023lidar,sima2023drivelm}.

\begin{table*}
\centering
\renewcommand{\arraystretch}{1.5} 
\caption{Overview of Frameworks to test LLMs for AD (see \cref{sec:EvaluationBenchmark})}
\vspace{-1em} 
\label{table:Evaluation}
\scriptsize
\begin{tabularx}{\textwidth}{lccX}
\hline
\textbf{Model} & \textbf{Year} & \textbf{Task} & \multicolumn{1}{c}{\textbf{Model Description}} \\ \hline
LingoQA \cite{marcu2023lingoqa} & 2023 & VQA, Reasoning & Video question answering dataset and benchmark for autonomous driving. \\
Reason2Drive \cite{nie2023reason2drive} & 2023 & VQA, Reasoning & Benchmark dataset with video-text pairs designed for the reasoning in complex driving scenarios. \\
LiDAR-LLM \cite{yang2023lidar} & 2023 & VQA, Scene Und.,  & Model interprets raw LiDAR data through the language modeling to understand outdoor 3D scenes. \\
Nuscenes-QA \cite{qian2024nuscenes} & 2023 & VQA  &  VQA benchmark for AD that features 34K visual scenes and 460K question-answer pairs. \\
DRAMA \cite{malla2023drama} & 2023 & VQA, Reasoning &  Large-scale collection of interactive driving scenarios from Tokyo. \\
DriveLM \cite{sima2023drivelm} & 2023 & VQA, Reasoning, Scene Und.,  & Dataset and benchmark based on nuScenes and CARLA. \\
VLAAD \cite{park2024vlaad} & 2024 & VQA, Reasoning & Dataset for interpreting visual instructions in diverse driving scenarios. \\
NuScenes-MQA \cite{inoue2024nuscenes} & 2024 & VQA & Dataset to test the model’s capabilities in sentence generation and VQA. \\
\end{tabularx}
\vspace{-2em} 
\end{table*}

\subsubsection{Planning}
\label{sec:Planning}


In the planning phase, LLMs can replace human drivers by making end-to-end decisions autonomously. There are two main paradigms: fine-tuning pre-trained language models for AD \cite{mao2023gpt, xu2023drivegpt4} and using language models to drive cars based on human prompts \cite{cui2024drive}.



\begin{table*}
\centering
\renewcommand{\arraystretch}{1.5} 
\caption{An overview of generative models for VQA and Planning in AD (see \cref{sec:QuestionAnswering} and \cref{sec:Planning} )}
\vspace{-1em} 
\label{table:VQAandPlanning}
\scriptsize
\begin{tabularx}{\textwidth}{ccccX}
\hline
\textbf{Model} & \textbf{Year} & \textbf{Task} & \textbf{Input Format} & \multicolumn{1}{c}{\textbf{Model Description}} \\ \hline
DriveMLM \cite{wang2023drivemlm} & 2023 & VQA-Planning & Image-Point Cloud-Prompt &  Interprets both point clouds and text prompts to generate control decisions.   \\
DriveGPT4 \cite{xu2023drivegpt4} & 2023 & VQA-Planning  & Image-Video  &  For traffic scene reasoning, capable of real-time responses and decision explanations. \\
DriveLM \cite{sima2023drivelm} & 2023 & VQA & Video-Prompt & Leverages a graph-based approach to sequence question-answer pairs in a logical order  \\
Dolphins \cite{ma2023dolphins} & 2023 & VQA & Image-Prompt & Grounded chain of thought process to improve driving performance. \\
LMDrive \cite{shao2023lmdrive} & 2023 & Planning & Image-Lidar-Prompt & End-to-end AD framework that utilizes sensor data and natural language for driving. \\
Lingo-1 \cite{wayve_lingo} & 2023 & VQA-Planning & Image-Prompt & End-to-end AD framework that utilizes sensor data and natural language for driving. \\
Surreal Driver \cite{jin2023surrealdriver}  & 2023 & Planning   & Image-Prompt   & Driver simulation framework using LLM to create traffic scenarios and maneuvers \\
Agent-Driver \cite{mao2023language} & 2023 & VQA-Planning & Image-Prompt & LLM as an agent for AD with cognitive memory and reasoning engine. \\
DriveVLM \cite{tian2024drivevlm} & 2024 & VQA-Planning & Image-Prompt & Chain-of-thought modules for scene understanding and planning. \\
DMEDriver \cite{han2024dme} & 2024 & VQA-Planning & Image-Prompt & VLM for decision-making and perception for control and environmental perception. \\ 
ELM \cite{zhou2024embodied} & 2024 & VQA-Planning & Image-Prompt & Driving agents with space-aware training for spatial and temporal perception. \\
Lingo-2 \cite{wayve_lingo2_2024} & 2024 & VQA-Planning & Image-Prompt & End-to-end AD framework that utilizes sensor data and natural language for driving. \\
RAG-Driver \cite{yuan2024rag} & 2024 & VQA & Image-Prompt & Retrieval-augmented multi-modal LLM, to enhance explainability in AD. \\
BEV-MLLM \cite{ding2024holistic} & 2024 & VQA-Planning & Image-Prompt & A module for LLMs, combining multiview Bird’s-Eye-View features. \\
VLP \cite{pan2024vlp} & 2024 & Planning & Image & Framework to integrate LLM for AD by improving memory and context understanding. \\
LLaDA \cite{li2024driving} & 2024 & Planning & Image & Tool to adapt driving behavior and motion plans to local traffic rules. \\
Lego-Drive \cite{paul2024lego} & 2024 & Planning & Image-Prompt & Model that estimates goal locations from language commands. \\
DriveasYouSpeak \cite{cui2024drive} & 2024  & Planning  & Image-Lidar-Radar-Prompt & Human like interactions with LLM in the driving task\\
\end{tabularx}
\vspace{-2em} 
\end{table*}

\section{Perception Safety in EU AI Act}
\label{sec:euaiact}


The EU AI Act \cite{euaiactcompromise2024} aims to ensure AI systems in the EU are trustworthy, safe, and beneficial while minimizing negative impacts. This survey focuses on perception safety in AD, identifying three key challenges related to the Act's requirements\footnote{Based on the final draft released in 2024 \cite{euaiactcompromise2024}}. We discuss these challenges and how generative AI could address them.


\subsection{Transparency}
\label{sec:Transparency}

Transparency in AI systems, particularly for high-risk applications like AD, is paramount for ensuring safety, trust, and regulatory compliance. According to the EU AI Act, transparency encompasses the development and use of AI systems in a manner that allows for appropriate traceability and explainability. This includes: (1) \textit{Making humans aware that they are interacting with an AI system.} (2) \textit{Informing deployers about the AI system's capabilities and limitations.} (3) \textit{Notifying affected individuals about their rights.} For high-risk AI systems, such as those used in AD, the Act emphasizes transparency regarding model capabilities and limitations.

DNNs, commonly used in AD perception, posing significant challenges in verifying and interpreting their decisions, understanding model limitations, and identifying potential failure cases. Various explainability methods exist for DNNs in AD perception, but they have notable limitations: some methods are model-specific \cite{muhammad2020deep}, others provide only post-hoc explanations \cite{keser2023interpretable}, and many are not human-interpretable \cite{atakishiyev2024safety}. These limitations hinder the ability to fully meet the transparency requirements mandated by the EU AI Act.

Multimodal LLMs introduced in \cref{sec:GenModelsDataDecisionMaking} offer a promising solution to address transparency requirements by providing reasoned, easily understandable explanations for their decisions. These human-interpretable explanations enhance user trust and comprehension. Furthermore, these models possess interactive capabilities, allowing users to query them about detections and decision-making processes. This interaction facilitates better understanding and engagement with the system. Additionally, multimodal LLMs support flexible communication timeframes, offering both immediate feedback within 0-1 seconds and post-action behavioral explanations. To comply with the EU AI Act using multimodal LLMs, developers could leverage these features to ensure that their systems are transparent, user-friendly, and adaptable to various regulatory requirements.

To comply with the EU AI Act using multimodal LLMs, developers could:
\vspace{-0.3em} 
\begin{itemize}[left=0pt, partopsep=0pt]
\item \textbf{Real-time Explanation:} Implement a system where the LLM provides immediate, natural language explanations for critical decisions made by the AD system.
\item \textbf{User Interaction:} Develop an interface allowing users to ask questions about the system's perceptions and decisions, with the LLM providing context-aware responses.
\item \textbf{Capability Disclosure:} Use the LLM to generate comprehensive, understandable descriptions of the AI system's capabilities and limitations for deployers and users.
\item \textbf{Traceability Logs:} Employ the LLM to maintain and interpret detailed logs of the AI system's decision-making processes, enhancing traceability.
\end{itemize}
By integrating these approaches, developers can create more transparent AD systems that align with the EU AI Act's requirements while enhancing user trust and system safety.


\vspace{-0.3em} 
\subsection{Accuracy \& Robustness}
\label{sec:Robustness}

Accuracy and robustness are essential for the safe operation of AD systems, as required by the EU AI Act. These systems must perform reliably across diverse environments, including varying weather and unexpected obstacles, which introduces significant challenges. DNNs, central to these systems, often struggle to generalize to new scenarios, leading to potential safety-critical errors. Additionally, the black-box nature of DNNs complicates interpretation and debugging, making it difficult to enhance performance and ensure robustness in all conditions.

Several strategies have been employed to improve the accuracy and robustness of DNN-based AD systems, including \textit{data augmentation}, \textit{ensemble methods}, and \textit{adversarial training}. \textit{Data augmentation} uses techniques like synthetic data generation and domain randomization to expose models to a wider variety of scenarios during training, enhancing their generalization ability. \textit{Ensemble methods} combine multiple models to mitigate individual weaknesses and achieve more robust performance. \textit{Adversarial training} involves training models with adversarial examples to enhance their resilience to unexpected inputs and perturbations.

While traditional methods have significantly contributed to the accuracy and robustness of AD systems, generative AI models, that are listed in the \cref{table:World_Models,table:Simulation_Methods}, offer additional capabilities that further enhance. Generative models can create diverse and realistic driving scenarios that go beyond traditional data augmentation, providing a richer variety of training data, including rare and challenging situations. This scenario generation helps train and test autonomous systems to perform accurately across a wide range of conditions. Furthermore, by generating synthetic data, generative models offer advanced data augmentation, supplementing real-world data to enhance the robustness of perception systems and improve their ability to generalize. World models also play a crucial role by simulating entire environments, including dynamic and static objects, and predicting future states, thereby creating photorealistic and physically accurate scenarios. This approach allows for improved training and testing of AD systems, offering fidelity and predictive capabilities often lacking in traditional methods.

\begin{figure}[ht]
\vspace{-0.5em} 
\centering
\includegraphics[width=1\columnwidth]{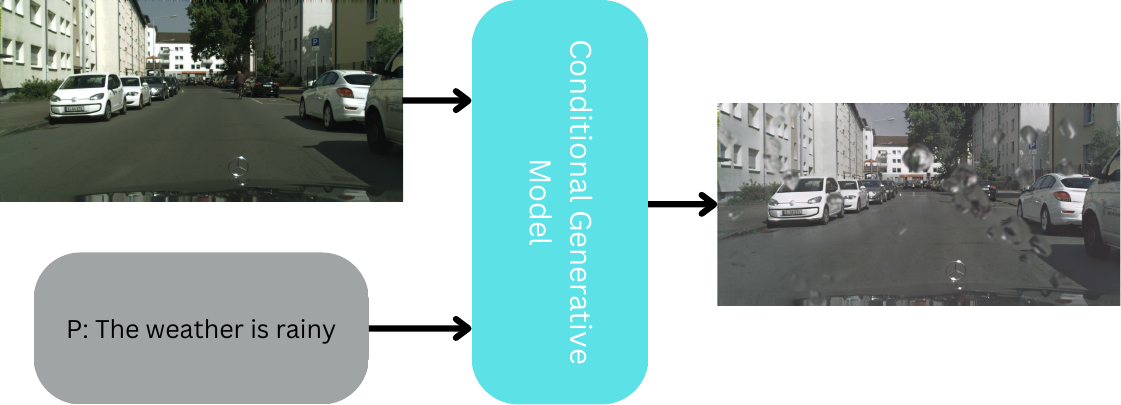}
\vspace{-2em} 
\caption{Generating diverse driving scenarios with, \\ e.g., DIDEX \cite{niemeijer2024generalization}}
\label{fig:ConditionalImageGeneration}
\vspace{-1em} 
\end{figure}

To align with the EU AI Act's requirements for accuracy and robustness, developers can implement generative AI models in the following ways:
\vspace{-0.3em} 
\begin{itemize}[left=0pt, partopsep=0pt]
    \item \textbf{Comprehensive Scenario Libraries:} Develop extensive libraries of generated scenarios that cover a wide range of environmental conditions and edge cases. These libraries can be used to rigorously test and validate AD systems.
    \item \textbf{Continuous Learning:} Implement systems that continuously update and refine their models based on new data and scenarios generated by generative AI. This approach ensures that the models remain accurate and robust as they encounter new situations.
    \item \textbf{Adversarial Testing:} Use generative models to create adversarial examples that test the AD system. This method helps to identify and address potential vulnerabilities.
    \item \textbf{Simulation-Based Validation:} Leverage world models to simulate real-world driving conditions and validate the performance of AD systems in a controlled environment. 
\end{itemize}
By integrating these strategies, developers can enhance the accuracy and robustness of AD systems, ensuring they meet the stringent requirements of the EU AI Act while maintaining high levels of safety and reliability. An example of driving scene generation with different conditions is illustrated in Figure \ref{fig:ConditionalImageGeneration}.

\subsection{Run-Time Monitoring}
\label{sec:Monitoring}


Run-time monitoring is crucial for maintaining the safety and reliability of AD systems, traditionally involving several key methods. \textit{Health monitoring} is one key technique, continuously assessing system components to detect faults or anomalies in real-time. \textit{Sensor fusion} integrates data from multiple sensors to improve the accuracy and reliability of environmental perception. Additionally, \textit{redundancy} and \textit{fail-safe mechanisms} ensure that vehicles can safely handle unexpected failures by providing backup systems and protocols. Collectively, these traditional strategies offer a comprehensive approach to ensuring the ongoing safety and performance of autonomous vehicles.

Traditional methods provide a robust framework for detecting and responding to issues as they arise, ensuring that AD systems operate safely under a variety of conditions. While these methods are effective, generative AI models offer additional capabilities that significantly enhance run-time monitoring. Specifically, generative models improve anomaly detection, predictive maintenance, and dynamic scenario simulation in AD systems. By learning normal operational patterns, generative models can detect deviations that indicate potential faults, allowing for precise and early issue detection. Through the analysis of historical data and prediction of future states, these models can forecast potential failures, enabling proactive maintenance and reducing downtime. Moreover, generative models simulate various driving scenarios in real-time, allowing the system to anticipate and prepare for potential hazards, which enhances its ability to respond effectively to unexpected situations.

To leverage the full potential of generative AI for run-time monitoring, developers can implement the following strategies:

\begin{itemize}[left=0pt, partopsep=0pt]
    \item \textbf{Real-Time Anomaly Detection:} Utilize generative models to continuously monitor system performance and detect anomalies in real-time. This approach ensures that potential issues are identified and addressed promptly.
    \item \textbf{Scenario-Based Testing:} Use generative models to simulate a wide range of driving scenarios during run-time. This real-time simulation allows the system to anticipate and react to potential hazards, improving overall safety.
    \item \textbf{Adaptive Learning:} Develop systems that continuously learn and adapt based on new data and scenarios generated by generative AI. This continuous learning process ensures that the system remains robust and accurate over time.
    \item \textbf{Integrated Monitoring Dashboards:} Create comprehensive monitoring dashboards that integrate data from generative models, providing operators with real-time insights into system performance and potential issues.
\end{itemize}

\begin{figure}[ht]
\vspace{-1em} 
\centering
\includegraphics[width=\columnwidth]{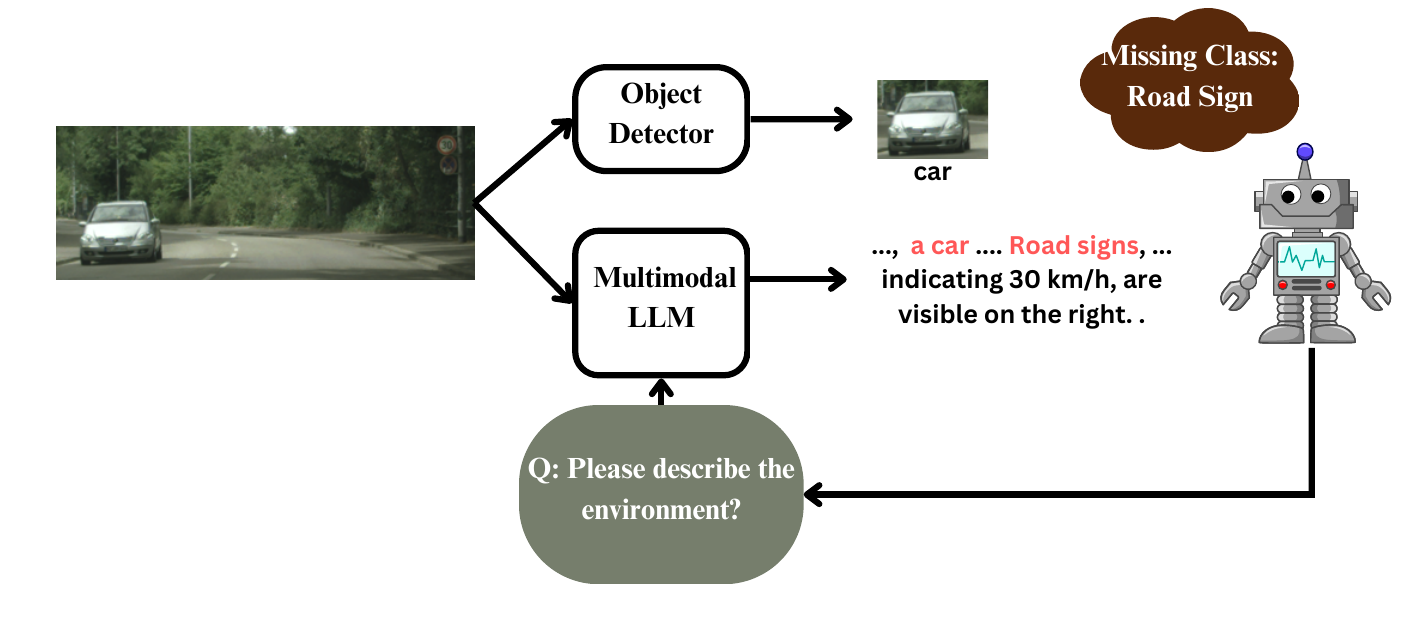}
\vspace{-2em} 
\caption{The event finder flowchart shows that multimodal LLMs can be utilized to identify events during the deployment. After the deployment, the object detector can be retrained, or the event can be recorded.}
\label{fig:EventFinder}
\vspace{-1.5em} 
\end{figure}

By integrating these generative AI strategies, developers can significantly enhance the run-time monitoring capabilities of AD systems. This approach not only ensures compliance with the stringent requirements of the EU AI Act but also improves the overall safety, reliability, and performance of the system. Figure \ref{fig:EventFinder} shows an example of using LLMs for real-time monitoring of object detectors.

\section{Discussion}
\label{sec:discussion}

Recent advancements in generative AI have led to a diverse range of applications in AD. These applications include data generation methods, world models, and multi-modal LLMs. Multi-modal LLMs demonstrate capabilities in chain-of-thought and visual reasoning, essential for AD tasks. Moreover, these capabilities are crucial for enhancing AI safety in AD perception and meeting legislative requirements such as EU AI Act. Unlike traditional detectors trained on closed-set datasets, the visual reasoning and free-form text descriptions offered by these models provide richer semantic information. This can address long-tail detection challenges, such as the identification of novel classes and the interpretation of police hand signals. Multi-modal foundational models also show strong generalization abilities in handling various long-tail scenarios. Additionally, multi-modal LLMs can serve as end-to-end planners in AD. However, there are still limitations to both multi-modal LLMs and generative AI in this context.

\textbf{Hallucination Problem}: The hallucination of LLMs may have a negative impact on their applications. This hallucination issue disables the use of multimodal LLMs in the safety-critical systems. However, with the advancements in LLMs, the concern regarding hallucination is expected to be alleviated in the future.

\textbf{Temporal Scene Understanding}: Video data presents significant challenges for understanding temporal scenes due to its size and complexity compared to images or text. Processing long sequences of frames is essential to capture context for tasks like driving scenario comprehension. Determining causes and relationships, such as identifying the root cause of a traffic accident from video footage, can be complex. Thus, novel approaches are needed to effectively process and extract insights from high-dimensional temporal data.


\textbf{Computational Resource Availability}: In AD, real-time processing and limited computational resources pose significant challenges. Deploying generative models in perception systems requires deciding whether to place them on the vehicle or in the cloud, balancing real-time processing with access to computational resources. Solutions include on-demand execution, periodic checking, and hybrid edge-cloud collaboration, each offering unique benefits for managing resources and ensuring real-time responsiveness.

\section{CONCLUSIONS}
\label{sec:conclusion}

This survey has reviewed recent advancements in generative AI and their implications for enhancing perception safety in AD, particularly concerning compliance with the EU AI Act. Our analysis highlights how generative AI can address the Act's requirements for transparency, accuracy, and run-time monitoring, thereby improving the safety and reliability of perception of autonomous systems. By leveraging generative AI, developers can create more robust AD models that meet legislative standards while adapting to evolving technological challenges.







\bibliographystyle{plain}
\bibliography{main} 

\begin{thebibliography}{10}

\bibitem{meta-llama3-webpage}
Meta AI.
\newblock Meta llama 3, 2024.
\newblock Accessed on [04.29.2024].

\bibitem{euaiactcompromise2024}
{Council of the European Union}.
\newblock Artificial intelligence act - final compromise text.
\newblock \url{https://data.consilium.europa.eu/doc/document/ST-5662-2024-INIT/en/pdf}, 2024.
\newblock Accessed: 2024-03-23.

\bibitem{marathe2023wedge}
Aboli~Marathe et~al.
\newblock Wedge: A multi-weather autonomous driving dataset built from generative vision-language models.
\newblock In {\em CVPR}, 2023.

\bibitem{ramesh2022hierarchical}
Aditya~Ramesh et~al.
\newblock Hierarchical text-conditional image generation with clip latents.
\newblock {\em arXiv preprint arXiv:2204.06125}, 2022.

\bibitem{marcu2023lingoqa}
Ana-Maria~Marcu et~al.
\newblock Lingoqa: Video question answering for autonomous driving.
\newblock {\em arXiv:2312.14115}, 2023.

\bibitem{hu2022model}
Anthony~Hu et~al.
\newblock Model-based imitation learning for urban driving.
\newblock {\em Advances in Neural Information Processing Systems}, 2022.

\bibitem{hu2023gaia}
Anthony~Hu et~al.
\newblock Gaia-1: A generative world model for autonomous driving.
\newblock {\em arXiv:2309.17080}, 2023.

\bibitem{vaswani2017attention}
Ashish~Vaswani et~al.
\newblock Attention is all you need.
\newblock {\em Advances in Neural Information Processing Systems}, 30, 2017.

\bibitem{li2024driving}
Boyi~Li et~al.
\newblock Driving everywhere with large language model policy adaptation.
\newblock {\em arXiv:2402.05932}, 2024.

\bibitem{cui2024drive}
Can~Cui et~al.
\newblock Drive as you speak: Enabling human-like interaction with large language models in autonomous vehicles.
\newblock In {\em WACV}, 2024.

\bibitem{cui2024survey}
Can~Cui et~al.
\newblock A survey on multimodal large language models for autonomous driving.
\newblock In {\em WACV}, 2024.

\bibitem{min2023uniworld}
Chen~Min et~al.
\newblock Uniworld: Autonomous driving pre-training via world models.
\newblock {\em arXiv:2308.07234}, 2023.

\bibitem{pan2024vlp}
Chenbin~Pan et~al.
\newblock Vlp: Vision language planning for autonomous driving.
\newblock {\em arXiv:2401.05577}, 2024.

\bibitem{sima2023drivelm}
Chonghao~Sima et~al.
\newblock Drivelm: Driving with graph visual question answering.
\newblock {\em arXiv:2312.14150}, 2023.

\bibitem{bogdoll2023muvo}
Daniel~Bogdoll et~al.
\newblock Muvo: A multimodal generative world model for autonomous driving.
\newblock {\em arXiv:2311.11762}, 2023.

\bibitem{fu2024drive}
Daocheng~Fu et~al.
\newblock Drive like a human: Rethinking autonomous driving with large language models.
\newblock In {\em WACV}, 2024.

\bibitem{kingma2013auto}
Diederik P~Kingma et~al.
\newblock Auto-encoding variational bayes.
\newblock {\em arXiv:1312.6114}, 2013.

\bibitem{yurtsever2020survey}
Ekim~Yurtsever et~al.
\newblock A survey of autonomous driving: Common practices and emerging technologies.
\newblock {\em IEEE Access}, 8:58443--58469, 2020.

\bibitem{jia2023adriver}
Fan~Jia et~al.
\newblock Adriver-i: A general world model for autonomous driving.
\newblock {\em arXiv:2311.13549}, 2023.

\bibitem{velasco2020autonomous}
Gustavo Velasco-Hernandez et~al.
\newblock Autonomous driving architectures, perception and data fusion: A review.
\newblock In {\em ICCP}, 2020.

\bibitem{shao2023lmdrive}
Hao~Shao et~al.
\newblock Lmdrive: End-to-end driving with large language models.
\newblock {\em arXiv:2312.07488}, 2023.

\bibitem{goodfellow2020generative}
Ian~Goodfellow et~al.
\newblock Generative adversarial networks.
\newblock {\em Communications of the ACM}, 63(11):139--144, 2020.

\bibitem{sutskever2011generating}
Ilya~Sutskever et~al.
\newblock Generating text with recurrent neural networks.
\newblock In {\em Proc. 28th Int. Conf. Machine Learning (ICML-11)}, pages 1017--1024, 2011.

\bibitem{mao2023gpt}
Jiageng~Mao et~al.
\newblock Gpt-driver: Learning to drive with gpt.
\newblock {\em arXiv:2310.01415}, 2023.

\bibitem{mao2023language}
Jiageng~Mao et~al.
\newblock A language agent for autonomous driving.
\newblock {\em arXiv:2311.10813}, 2023.

\bibitem{liudiffusion}
Jiahui~Liu et~al.
\newblock Diffusion-based data generation for out-of-distribution object detection.

\bibitem{yuan2024rag}
Jianhao~Yuan et~al.
\newblock Rag-driver: Driving explanations with retrieval-augmented in-context learning in large language models.
\newblock {\em arXiv:2402.10828}, 2024.

\bibitem{niemeijer2024generalization}
Joshua~Niemeijer et~al.
\newblock Generalization by adaptation: Diffusion-based domain extension for semantic segmentation.
\newblock In {\em WACV}, 2024.

\bibitem{muhammad2020deep}
Khan~Muhammad et~al.
\newblock Deep learning for safe autonomous driving: Current challenges and future directions.
\newblock {\em IEEE Trans. Intelligent Transportation Systems}, 2020.

\bibitem{yang2023diffusion}
Ling~Yang et~al.
\newblock Diffusion models: A comprehensive survey of methods and applications.
\newblock {\em ACM Computing Surveys}, 2023.

\bibitem{keser2023interpretable}
Mert~Keser et~al.
\newblock Interpretable model-agnostic plausibility verification for 2d object detectors using domain-invariant concept bottleneck models.
\newblock In {\em CVPR}, pages 3891--3900, 2023.

\bibitem{nie2023reason2drive}
Ming~Nie et~al.
\newblock Reason2drive: Towards interpretable and chain-based reasoning for autonomous driving.
\newblock {\em arXiv:2312.03661}, 2023.

\bibitem{paul2024lego}
Pranjal~Paul et~al.
\newblock Lego-drive: Language-enhanced goal-oriented closed-loop autonomous driving.
\newblock {\em arXiv:2403.20116}, 2024.

\bibitem{qian2024nuscenes}
Qian~Tianwen et~al.
\newblock Nuscenes-qa: Visual question answering for autonomous driving.
\newblock In {\em Proc. AAAI Conf. Artifical Intell.}, pages 4542--4550, 2024.

\bibitem{li2024think2drive}
Qifeng~Li et~al.
\newblock Think2drive: Efficient reinforcement learning in latent world model for autonomous driving.
\newblock {\em arXiv:2402.16720}, 2024.

\bibitem{Rombach_2022_CVPR}
Robin~Rombach et~al.
\newblock High-resolution image synthesis with latent diffusion models.
\newblock In {\em CVPR}, 2022.

\bibitem{gong2023prompting}
Rui~Gong et~al.
\newblock Prompting diffusion representations for cross-domain semantic segmentation.
\newblock {\em arXiv:2307.02138}, 2023.

\bibitem{gao2023magicdrive}
Ruiyuan~Gao et~al.
\newblock Magicdrive: Street view generation with diverse 3d geometry control.
\newblock {\em arXiv:2310.02601}, 2023.

\bibitem{yang2023lidar}
Senqiao~Yang et~al.
\newblock Lidar-llm: Exploring large language models for 3d lidar understanding.
\newblock {\em arXiv:2312.14074}, 2023.

\bibitem{hochreiter1997long}
Sepp~Hochreiter et~al.
\newblock Long short-term memory.
\newblock {\em Neural Computation}, 1997.

\bibitem{atakishiyev2024safety}
Shahin~Atakishiyev et~al.
\newblock Safety implications of explainable ai in end-to-end autonomous driving.
\newblock {\em arXiv:2403.12176}, 2024.

\bibitem{sun2023drivescenegen}
Shuo~Sun et~al.
\newblock Drivescenegen: Generating diverse and realistic driving scenarios from scratch.
\newblock {\em arXiv:2309.14685}, 2023.

\bibitem{grigorescu2020survey}
Sorin~Grigorescu et~al.
\newblock A survey of deep learning techniques for autonomous driving.
\newblock {\em J. Field Robotics}, 37(3):362--386, 2020.

\bibitem{malla2023drama}
Srikanth~Malla et~al.
\newblock Drama: Joint risk localization and captioning in driving.
\newblock In {\em WACV}, pages 1043--1052, 2023.

\bibitem{park2024vlaad}
SungYeon~Park et~al.
\newblock Vlaad: Vision and language assistant for autonomous driving.
\newblock In {\em WACV}, pages 980--987, 2024.

\bibitem{loiseau2023reliability}
Thibaut~Loiseau et~al.
\newblock Reliability in semantic segmentation: Can we use synthetic data?
\newblock {\em arXiv:2312.09231}, 2023.

\bibitem{ponn2020identification}
Thomas~Ponn et~al.
\newblock Identification and explanation of challenging conditions for camera-based object detection of automated vehicles.
\newblock {\em Sensors}, 20(13):3699, 2020.

\bibitem{videoworldsimulators2024}
Tim~Brooks et~al.
\newblock Video generation models as world simulators.
\newblock 2024.

\bibitem{brown2020language}
Tom~Brown et~al.
\newblock Language models are few-shot learners.
\newblock {\em Advances in Neural Information Processing Systems}, 2020.

\bibitem{boreiko2023identifying}
Valentyn~Boreiko et~al.
\newblock Identifying systematic errors in object detectors with the scrod pipeline.
\newblock In {\em CVPR}, 2023.

\bibitem{han2024dme}
Wencheng~Han et~al.
\newblock Dme-driver: Integrating human decision logic in autonomous driving.
\newblock {\em arXiv:2401.03641}, 2024.

\bibitem{wang2023drivemlm}
Wenhai~Wang et~al.
\newblock Drivemlm: Aligning multi-modal large language models with behavioral planning states for autonomous driving.
\newblock {\em arXiv:2312.09245}, 2023.

\bibitem{zheng2023occworld}
Wenzhao~Zheng et~al.
\newblock Occworld: Learning a 3d occupancy world model for autonomous driving.
\newblock {\em arXiv:2311.16038}, 2023.

\bibitem{li2023drivingdiffusion}
Xiaofan~Li et~al.
\newblock Drivingdiffusion: Layout-guided multi-view driving scene video generation with latent diffusion model.
\newblock {\em arXiv:2310.07771}, 2023.

\bibitem{wang2023drivedreamer}
Xiaofeng~Wang et~al.
\newblock Drivedreamer: Towards real-world-driven world models for autonomous driving.
\newblock {\em arXiv:2309.09777}, 2023.

\bibitem{wang2024worlddreamer}
Xiaofeng~Wang et~al.
\newblock Worlddreamer: Towards general world models for video generation via predicting masked tokens.
\newblock {\em arXiv:2401.09985}, 2024.

\bibitem{tian2024drivevlm}
Xiaoyu~Tian et~al.
\newblock Drivevlm: Autonomous driving and vision-language models.
\newblock {\em arXiv:2402.12289}, 2024.

\bibitem{zhou2023vision}
Xingcheng~Zhou et~al.
\newblock Vision language models in autonomous driving and intelligent transportation systems.
\newblock {\em arXiv:2310.14414}, 2023.

\bibitem{ding2024holistic}
Xinpeng~Ding et~al.
\newblock Holistic autonomous driving understanding with bird's-eye-view injected multi-modal models.
\newblock {\em arXiv:2401.00988}, 2024.

\bibitem{jin2023surrealdriver}
Ye~Jin et~al.
\newblock Surrealdriver: Generative driver agent simulation framework.
\newblock {\em arXiv:2309.13193}, 2023.

\bibitem{ma2023dolphins}
Yingzi~Ma et~al.
\newblock Dolphins: Multimodal language model for driving.
\newblock {\em arXiv:2312.00438}, 2023.

\bibitem{huang2023applications}
Yu~Huang et~al.
\newblock Applications of large scale foundation models for autonomous driving.
\newblock {\em arXiv preprint arXiv:2311.12144}, 2023.

\bibitem{inoue2024nuscenes}
Yuichi~Inoue et~al.
\newblock Nuscenes-mqa: Evaluation of captions and qa for autonomous driving datasets.
\newblock In {\em WACV}, 2024.

\bibitem{zhou2024embodied}
Yunsong~Zhou et~al.
\newblock Embodied understanding of driving scenarios.
\newblock {\em arXiv:2403.04593}, 2024.

\bibitem{wang2023driving}
Yuqi~Wang et~al.
\newblock Driving into the future: Multiview visual forecasting and planning with world model.
\newblock {\em arXiv:2311.17918}, 2023.

\bibitem{zhang2023trafficbots}
Zhejun~Zhang et~al.
\newblock Trafficbots: Towards world models for autonomous driving simulation and motion prediction.
\newblock {\em arXiv:2303.04116}, 2023.

\bibitem{xu2023drivegpt4}
Zhenhua~Xu et~al.
\newblock Drivegpt4: Interpretable end-to-end autonomous driving via large language model.
\newblock {\em arXiv:2310.01412}, 2023.

\bibitem{yang2023llm4drive}
Zhenjie~Yang et~al.
\newblock Llm4drive: A survey of large language models for autonomous driving.
\newblock {\em arXiv e-prints}, pages arXiv--2311, 2023.

\bibitem{guo2023controllable}
Zipeng~Guo et~al.
\newblock Controllable diffusion models for safety-critical driving scenario generation.
\newblock In {\em ICTAI}, pages 717--722. IEEE, 2023.

\bibitem{zhong2023language}
Ziyuan~Zhong et~al.
\newblock Language-guided traffic simulation via scene-level diffusion.
\newblock {\em arXiv:2306.06344}, 2023.

\bibitem{OpenAIChatGPT}
OpenAI.
\newblock Introducing chatgpt.
\newblock \url{https://openai.com/blog/chatgpt}, 2022.
\newblock Accessed: 2024-02-01.

\bibitem{wayve_lingo2_2024}
Wayve.
\newblock Lingo 2: Driving with language.
\newblock Accessed: May 2, 2024.

\bibitem{wayve_lingo}
{Wayve}.
\newblock Lingo: Natural language for autonomous driving, 2023.
\newblock Accessed: May 2, 2024.

\end{thebibliography}

\end{document}